\documentclass[dvipsnames]{article} 
\usepackage{amsmath}
\usepackage[preprint]{colm2025_conference}

\usepackage{microtype}
\usepackage{hyperref}
\usepackage{url}
\usepackage{booktabs}
\usepackage{natbib}
\usepackage{xcolor}
\usepackage{graphicx}
\usepackage{enumitem}
\usepackage{cleveref}
\usepackage{wrapfig}
\usepackage{multirow}
\usepackage{svg}
\usepackage{amsmath}
\usepackage{xcolor}
\usepackage{xcolor, soul}
\sethlcolor{pink}

\definecolor{darkblue}{rgb}{0, 0, 0.5}
\hypersetup{colorlinks=true, citecolor=darkblue, linkcolor=darkblue, urlcolor=darkblue}

\newcommand{\hypothesis}[1]{\texttt{\textbf{#1}}}

\definecolor{bluish}{HTML}{1f78b4}
\definecolor{greenish}{HTML}{1b9e77}
\definecolor{orangish}{HTML}{d95f02}

\newcommand{\zeroshot}{\textcolor{greenish}{\textbf{0-shot}}}
\newcommand{\cothought}{\textcolor{bluish}{\textbf{CoT}}}
\newcommand{\cotfull}{\textcolor{bluish}{\textbf{Chain-of-Thought}}}
\newcommand{\knowledge}{\textcolor{orangish}{\textbf{Knowledge}}}

\title{On Language Models' Sensitivity to Suspicious Coincidences}

\author{Sriram Padmanabhan$^{\diamondsuit}$, Kanishka Misra$^{\heartsuit}$\thanks{Work partly done at UT Austin before joining TTIC}, Kyle Mahowald$^{\diamondsuit}$, Eunsol Choi$^{\clubsuit}$\thanks{Work partly done at UT Austin before joining NYU} \\[2pt]
The University of Texas at Austin$^{\diamondsuit}$, Toyota Technological Institute at Chicago$^{\heartsuit}$, \\
New York University$^{\clubsuit}$ \\[2pt]
\texttt{srirampadmanabhan@utexas.edu, kanishka@ttic.edu, kyle@utexas.edu,} \\
\texttt{eunsol@nyu.edu} \\
}

\begin{document}

\ifcolmsubmission
% \linenumbers
\fi

\maketitle
\begin{abstract}
Humans are sensitive to suspicious coincidences when generalizing inductively over data, as they make assumptions as to how the data was sampled. This results in smaller, more specific hypotheses being favored over more general ones. For instance, when provided the set \{Austin, Dallas, Houston\}, one is more likely to think that this is sampled from ``Texas Cities'' over ``US Cities'' even though both are compatible. Suspicious coincidence is strongly connected to pragmatic reasoning, and can serve as a testbed to analyze systems on their sensitivity towards the communicative goals of the task (i.e., figuring out the true category underlying the data). 
In this paper, we analyze whether suspicious coincidence effects are reflected in language models' (LMs) behavior.
We do so in the context of two domains: 1) the number game, where humans made judgments of whether a number (e.g., 4) fits a list of given numbers (e.g., 16, 32, 2); and 2) by extending the number game setup to prominent cities. For both domains, the data is compatible with multiple hypotheses and we study which hypothesis is most consistent with the models' behavior. On analyzing five models, we do not find strong evidence for suspicious coincidences in LMs’ zero-shot behavior. However, when provided access to the hypotheses space via chain-of-thought or explicit prompting, LMs start to show an effect resembling suspicious coincidences, sometimes even showing effects consistent with humans. 
Our study suggests that inductive reasoning behavior in LMs can be enhanced with explicit access to the hypothesis landscape.

\end{abstract}

\section{Introduction}

When humans learn and generalize from data, they demonstrate sensitivities to the process by which the data is generated \citep{tenenbaum2001generalization}. Consider the Number Game \citep{tenenbaum1999Bayesian, tenenbaum2000rules} example: subjects are provided with a mysterious computer program that produces lists of integers between 1 and 100 according to some rule, and their task is to predict what numbers are likely to obey this rule. For instance, if the program produced \{2, 16, 4\}, then would it also produce `12'? How about `32' or `8'? In this setup, the input is often compatible with a range of different hypotheses---e.g., \hypothesis{natural numbers}, \hypothesis{even numbers}, \hypothesis{powers of 2}, etc. Among these multiple valid options, people tend to prefer specific hypotheses over general ones, i.e., favoring \hypothesis{powers of 2} over \hypothesis{even numbers} for our running example. This sensitivity to sampling of data has also been described as reasoning about \textit{suspicious coincidences}---it would be suspicious to observe only powers of two (2, 16, 4) as an example output from the program if it was using a more general rule like \hypothesis{even numbers}. 
Sensitivity to suspicious coincidence is prevalent and has been observed in a number of inductive reasoning problems \citep{tenenbaum2001generalization, raviv2022variability}, including early word learning \citep{xu2007word, xu2007sensitivity} and category-based induction \citep{osherson1990category, medin2003relevance, ransom2016leaping}.

Inspired by the literature on human inductive reasoning, we test whether sensitivities to suspicious coincidences arise in the reasoning behavior of language models (LMs). This can serve as a testbed to characterise LMs' ability to reason in human-like ways by taking advantage of structure in the data and in the nature of statistical sampling \citep{tenenbaum1999Bayesian,lake2017building}. Previous work that is relevantly close to ours is that of \citet{han2024inductive}, who report LMs' insensitivity to a phenomenon called ``premise non-monotonicity'', where under certain conditions, instead of the classical idea that more premise arguments in a reasoning domain lead to stronger conclusions, the opposite is observed \citep{osherson1990category, medin2003relevance}. For instance, humans judge the argument \textit{``all \textbf{animals} have a property''} to be weaker when told that \textit{``\textbf{polar bears}, \textbf{brown bears}, and \textbf{grizzly bears} have a property''} than when told that \textit{``\textbf{brown bears} have a property''}. The category of \textit{bears} is more salient in the former, making generalizations to \textit{animals} suspicious given the input. LMs were reported to struggle with this behavior when prompted only with the task instructions \citep{han2024inductive}. 
Can LMs demonstrate non-monotonic reasoning behaviors \citep{leidinger2024llms, han2024inductive}? Or unlike models that show suspicious coincidence effects \citep[e.g., Bayesian models, see][]{tenenbaum2001generalization, navarro2010similarity, frank2014inferring}, LMs might not be operating over a constrained hypothesis space? If so, does providing them access to the hypothesis landscape allow them to narrow in on the right hypothesis?

\begin{figure}[t]
    \centering
    \vspace{-1em}
    \includegraphics[width=\linewidth]{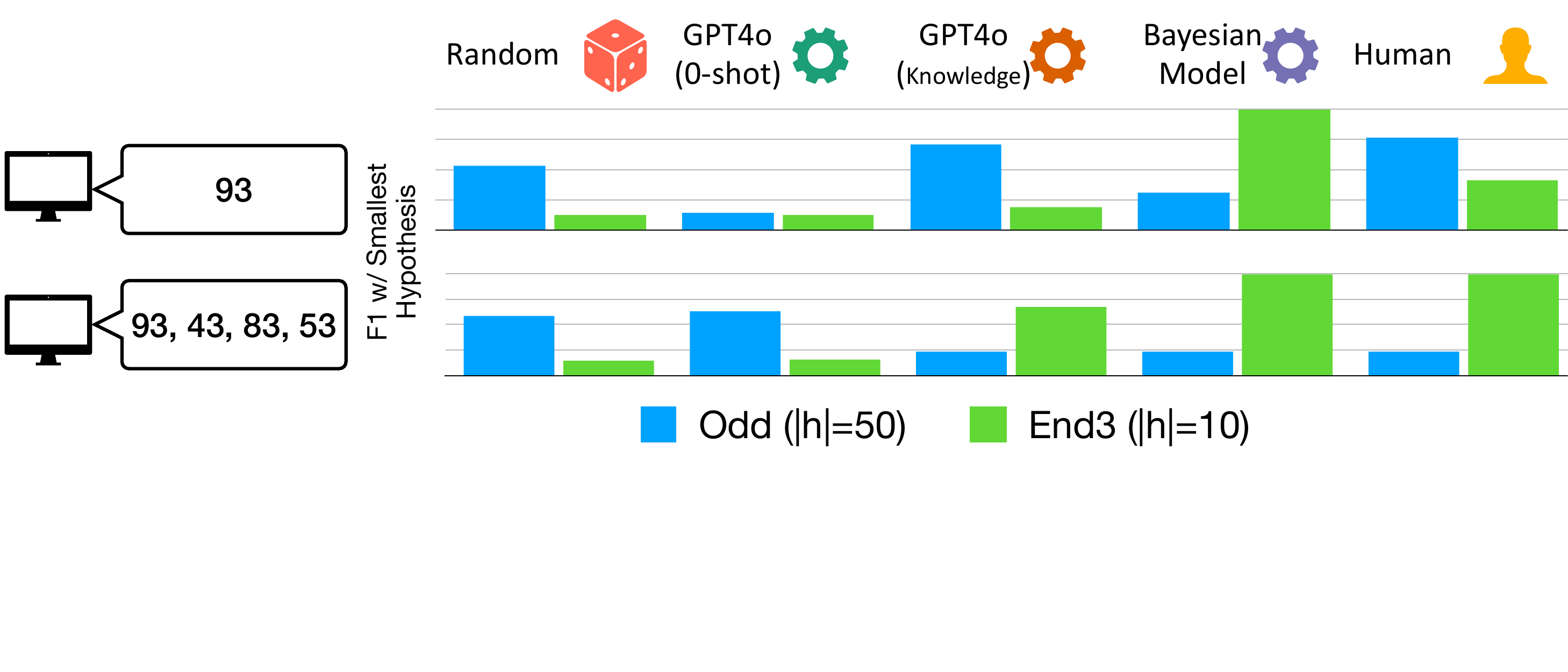}
    \vspace{-6.5em}
    \caption{Compatibility of predictions with two hypothesis (Odd, End3) from five systems: random baseline, GPT-4o (0-shot), GPT-4o (with access to hypotheses information, denoted as `Knowledge'), Bayesian model and human predictions. When suspicious coincidences is observed \{93, 43, 83, 53\}, three systems (Humans, Bayesian model, GPT-4o (Knowledge) strongly favors smaller hypothesis (End3), while other systems do not.}
    \label{fig:sus-coincidence}
    \vspace{-1em}
\end{figure}

To shed light on these questions, we test LMs using experimental paradigms that ask fine-grained questions about suspicious coincidences during concept learning in humans~\citep{tenenbaum1999Bayesian}. Specifically, we rely on a large-scale data set of human generalization patterns on the number game \citep{bigelow2016large}, and we evaluate LMs behavior on this domain as well as our newly devised domain of prominent cities. We analyze five competitive LMs (including Llama3-8b, GPT-4o). Like the human participants of~\citet{bigelow2016large}, LMs are provided with input examples that could have been generated with multiple hypotheses, and are tasked to predict if a query belongs to the set of input examples or not. We then measure the extent to which LMs' predictions are compatible with the smallest, most specific hypothesis that explains the input, and crucially, whether this compatibility increases as more input samples are provided. Figure~\ref{fig:sus-coincidence} visualizes our study.

We first evaluate LMs' sensitivity to suspicious coincidences across both domains with a 0-shot prompt (i.e., given an input, predict the membership of the query). Here we do not find evidence for suspicious coincidence effects in either domain. We hypothesize that this could either be due to LMs' not storing sufficient parametric knowledge relevant to the individual entities (numbers and cities) and the input examples, or it could be that LMs do actually have the prerequisite knowledge but do not necessarily reason over the same hypotheses that we have used as our lens for analysis. To adjudicate between these possibilities, we test whether LMs' knowledge about the hypothesis space can explain their lack of sensitivity to suspicious coincidences. We fail to find any such relation, ruling out the first hypothesis.

To test the second hypothesis, we run an experiment where LMs are guided to reason over the kinds of hypotheses we use in our analysis---either using chain of though prompting \citep{wei2022chain}, or by providing information about the hypothesis space in their input. The latter prompting method causes LMs to be more sensitive to suspicious coincidence effects---with GPT-4o even showing near-perfect alignment with human-behavior (in the case of the number game). This knowledge-based prompting method places LMs on level-ground with a Bayesian solution---i.e., both class of models have a more restricted set of candidate hypotheses to operate over (hypothesized in the case of LMs).
Overall, we take this as evidence that {LMs' sensitivity to salience in their inductive reasoning behavior crucially depends on access to the hypothesis landscape}. 

\section{Task, Data, Evaluation}

\subsection{The task, and the emergence of suspicious coincidence from Bayesian computation}
\label{sec:task-bayes}

Suspicious coincidences are encoded naturally in Bayesian models of cognition via the size principle \citep{tenenbaum2001generalization, navarro2010similarity}. Before we describe the size principle, we first lay out the premise of the task:

\paragraph{Task} Given a sequence of $n$ input examples $X = \{x_1, \dots, x_n\}$ that belong to a category $C$, the task is to produce a decision if a target example $y$ also belongs to $C$. This is equivalent to a categorization task, but is not always framed that way during psychological experimentation \citep[e.g., see][]{tenenbaum1999Bayesian}.

\paragraph{Hypothesis space} Especially important to a Bayesian solution to this task is maintaining a set of $m$ hypotheses $H = \{h_1, \dots, h_m\}$ compatible with the input example. For instance, if the input is \{10, 50, 80\}, then the potential hypotheses could be \hypothesis{even, divisible-by-5, ends-in-0}, etc. Apart from being explicated in Bayesian solutions, the hypothesis space also offers us an important lens to analyze LM behavior, especially in the context of which hypotheses their behavior is most compatible with. These hypotheses represent what the eventual target category $C$ could be.

\paragraph{The Bayesian Solution} Given the above task and hypothesis space, a Bayesian learner updates its beliefs about $C$ by using the Bayes rule (Eq. \ref{eq:bayes}), and then averages over all applicable hypotheses (Eq. \ref{eq:averaging}):

\noindent
\begin{minipage}{.49\linewidth}
\begin{equation}
\label{eq:bayes}
    p(h \mid X) \propto p(X \mid h)p(h)
\end{equation}
\end{minipage}
\begin{minipage}{.49\linewidth}
\begin{equation}
    \label{eq:averaging}
    p(y \in C \mid X) = \sum_{h: y\in h}p(h\mid X)
\end{equation}
\end{minipage}

\noindent 
That is, the \textit{posterior} $p(h \mid X)$ is modeled by considering the likelihood of the data according to each hypothesis $p(X\mid h)$, weighed by the hypothesis' \textit{prior} probability $p(h)$. The prior is often kept to be uniform \citep{tenenbaum1999Bayesian}, while the likelihood determines the assumptions that the learner makes about how the input $X$ was sampled. This is where the \textit{size principle} makes its appearance. Under the assumption that $X$ has been explictly sampled (uniformly) from the true category, the likelihood is given by:

\begin{equation*}
    p(X \mid h) = \begin{cases}\frac{1}{\left|h\right|^n}, \textit{ if } x_1, \dots, x_n \in h\\0, \textit{ otherwise}\end{cases}
\end{equation*}
This way, smaller and more specific hypotheses will be preferred over larger, more general ones.
This preference increases exponentially as more data are encountered (i.e., as $n$ increases).
For Numbers, this predicts that a learner will be more suspicious that the input \{2, 16, 4\} is from a \textit{broad} hypothesis---i.e., strongly favor the more narrow hypothesis \hypothesis{powers of 2} ($\left|h\right|=7$) over the broader \hypothesis{even numbers} ($\left|h\right|=50$)---than when the input is just $\{2\}$. Beyond Numbers, this method has been successfully used to model human generalization patterns in early word learning \citep{xu2007word} as well as property induction \citep{ransom2016leaping}, with appropriately designed priors. In our experiments, we will use this method as an idealized size-principled learner.

\subsection{Data}
We study two datasets, each focusing on a different domain: (1) a large dataset of human behavior on the number game \citep{bigelow2016large}; and (2) our extension of the number game to the domain of prominent cities---albeit without human data.

\paragraph{Number game} 
Our first dataset originates from \citet{tenenbaum1999Bayesian}. Here, human participants were provided input examples consisting of positive integers between 1 and 100 (i.e., $X$) that was generated by a computer program, and were then tasked to respond Yes/No if a target integer (i.e., $y$) will also be produced by the program. We use human judgment data collected by \citet{bigelow2016large}, which consists of 255 total input sets ranging from 1 to 4 elements, along with 100 queries for each set (integers between 1 and 100). The dataset consists of responses from 606 participants, and each input-query pair was judged by an average of 12 participants. For every pair of input-set and query, we take the majority vote of the human judgments (`Yes' vs. `No') as the final human decision. We use the same hypothesis space as \citet{tenenbaum1999Bayesian}, consiting of 33 different hypotheses based on properties of the numbers, such as \hypothesis{even, odd, prime, divisible by 5 (div5), power of 2 (pow2), ends with 6 (end6)}, etc. \Cref{tab:number-hyps} in the appendix shows the list of hypotheses.

\begin{table}[]
\small
    \centering
    \begin{tabular}{p{4.5cm}|p{8.5cm}}
    \toprule
          Number & City  \\ \midrule
          (\textcolor{blue}{\{16, 32\}}, \hl{pow2, div8, div4, even}) & (\textcolor{blue}{\{San Fransisco, Milwaukee, San Antonio\}}, \hl{north america, west-hemis, north-hemis, temperate, developed})\\
         \\
             (\textcolor{blue}{\{96, 90, 6, 42\}}, \hl{div6, div3, even}) & (\textcolor{blue}{\{Mumbai, Lagos, Jakarta, Guangzhou\}}, \hl{10M+, tropical, developing, east-hemis})\\ \bottomrule
    \end{tabular}
    \vspace{-0.5em}
    \caption{Examples (\textcolor{blue}{input}, \hl{matching hypotheses}) from two domains (Number, City). }\vspace{-1em}
    \label{tab:stat_example}
\end{table}
\paragraph{City game} 
Our second dataset uses the setup from the number game, and extends it to the domain of prominent cities of the world, which are likely to be plentiful in LMs' training data (via news articles, for instance). Like numbers, cities are entities that satisfy multiple properties at the same time---e.g., Montreal is a city in Canada, but it is also a city in the northern hemisphere, and also a city whose name starts with an `M'. This allows us to also investigate sensitivity to suspicious coincidences in two different domains where models' knowledge could also differ---it is possible LMs might store more accurate parametric knowledge about unambiguous cities than specific mathematical properties of numbers.

\begin{wrapfigure}{R}{0.38\textwidth}

\vspace{-2em}
\begin{tabular}{lrr}

\toprule
& Number & City\\ \midrule
         \# elements  & 100 & 500 \\
         \# hypothesis & 33 & 18\\
         \# input sets&146& 181 \\ \bottomrule
\end{tabular}
\caption{Data statistics.}\label{tab:data_stats}
  \centering
  \includegraphics[scale=0.5]{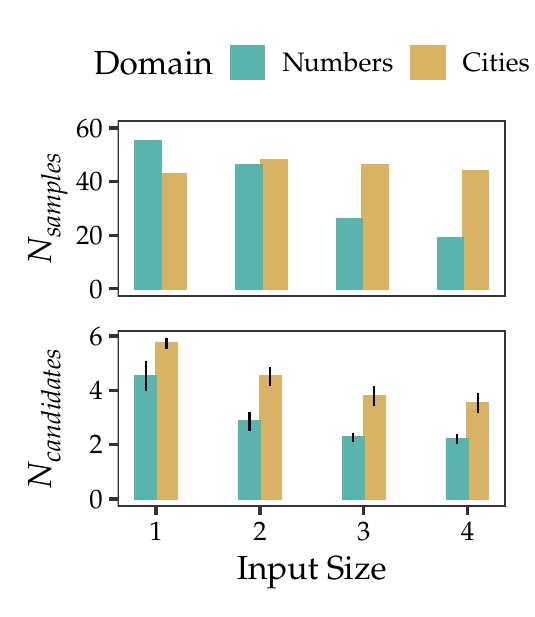}
\vspace{-2pt}
  \caption{The distribution of the number of input sets ($N_{samples}$; top plot) and the average number of competing hypotheses ($N_{candidates}$; bottom plot) across input set sizes.} \label{fig:avg_number_count}
\vspace{-4em}
\end{wrapfigure}

\begin{itemize}[leftmargin=10px]
    \item \textbf{{Elements}}
We start with data from Simplemaps,\footnote{\url{https://simplemaps.com/data/world-cities}} a comprehensive database of 48K world cities. Then, we approximate their prominence with their infini-gram score \citep{liu2024infinigramscalingunboundedngram}, which captures how frequently they occur in web corpora, and take 500 cities with highest scores. 
We manually excluded cities with ambiguous names—those overlapping with common words (e.g., Nice), countries (e.g., Luxembourg), similarly relevant cities (e.g., Newcastle), or common personal names (e.g., Victoria)—and manually replaced them with cities from underrepresented hypothesis categories (e.g., African cities). 
    \item{\textbf{{Hypotheses}} We then develop 18 different hypotheses that are based on the cities' geographical and demographic attributes---e.g., \hypothesis{african-cities, cities-that-are-capitals, cities-with-NFL-teams}, etc. The full list of hypotheses for the number and city domains are given in Appendix \ref{appendix:hypotheses}. The initial hypotheses included hypotheses such as \hypothesis{cities-that-hosted-summer-olympics}, which we pruned out as most LLMs cannot guess which cities hosted summer olympics reliably. }
    \item{\textbf{{Input Sets}} We manually create $50$ input sets of $4$ cities $[A, B, C, D]$. Then, for each such input, we add the incremental subsets $[A], [A, B], [A, B, C]$ in order to measure how model sensitivity to suspicious coincidence changes with input size. Note that this property of incremental subsets is absent from the number game data \citep{bigelow2016large}.}
\end{itemize}

\paragraph{Dataset Statistics}

Table~\ref{tab:stat_example} shows example instances and Figure~\ref{tab:data_stats} reports dataset statistics. 
From both domains, we filter any input that is only compatible with only one hypothesis, as our analyses require at least 2 competing hypotheses for there to be a notion of a `smaller' hypothesis. This gives us 146 and 181 final input sets for the number game and city game, respectively. \Cref{fig:avg_number_count} shows example instances from the two domains as well as statistics showing total number of instances per input-size. The city game data is more uniform across different input sizes. On average, data in both domains is consistent with 2--6 hypotheses.

\subsection{Measurement and Evaluation}

We investigate suspicious coincidences in LMs by analyzing the extent to which their judgments for the input-query pairs in our datasets is compatible with the smallest, most specific hypothesis applicable to the input. For instance, if the input set is \{Frankfurt, Marseille\}, the applicable hypotheses are ``european cities'' (\hypothesis{europe}), ``northern hemisphere'' (\hypothesis{north}), ``cities in developed countries'' (\hypothesis{developed}), etc., with \hypothesis{europe} being the smallest and most specific. Therefore, if the model systematically assigns `No' to cities like New York or Tokyo, and `Yes' to cities like Paris and Barcelona, then its behavior has greater compatibility with \hypothesis{europe} than with \hypothesis{north} or \hypothesis{developed}.

We measure compatibility with the smallest hypotheses by computing the F1 score using the LM responses (Yes/No) and the category-membership of all compatible hypotheses.
All our hypotheses are defined as \textit{categories} to which each query $y$ either belongs or does not. Therefore, each hypothesis can be associated with a binary vector of `yes' or `no' depending on the category membership of the queries in our dataset. Therefore, if an LM's predictions strongly match a particular hypothesis, then the F1 between them would be high. We report the F1 score here since many hypotheses are sparse (e.g., there are only 7 powers of 2 between 1 and 100), making the comparison equivalent to an imbalanced classification evaluation, where F1 is preferred.

In our analysis, we will report two values (derived from the above measure): (1) the percent of time the smallest hypothesis was preferred among all compatible hypotheses; and (2) the F1 score between the smallest hypothesis and the system predictions. The former will provide comparisons with other hypotheses, while the latter will provide absolute agreement with the smallest hypothesis. 
The higher these values get with increased input size, the greater the system's sensitivity to suspicious coincidences. 

\subsection{Comparison Systems} 
We construct two simple baselines: (1) a ``Random'' baseline for which we randomly sample Yes/No values 30 times for all input sets (to account for variability) and then select the majority label as the final prediction (2) an ``AlwaysYes'' baseline which only predicts `Yes' for every input-query pair. In addition to the baselines, we used the Bayesian model described in \Cref{sec:task-bayes} \citep[and first used by][]{tenenbaum1999Bayesian}. Here we took the model's response as Yes if the value of the term $p(y \in C \mid X)$ was greater than or equal to 0.5. Since this model explicitly implements the size-principle, it should always prefer the smallest hypothesis. For the numbers domain, we will also report the majority human judgements from prior work~\citep{bigelow2016large}. Our code and data can be found at https://github.com/kanishkamisra/number-game.

\section{Deriving Judgements from Language Models}\label{sec:models}

\paragraph{LLMs}We analyze three open-sourced, instruction-tuned LMs: \texttt{llama-3-8b-instruct} \citep{llama3}, \texttt{mistral-7b-instruct-v0.3} \citep{mistral7b}, and \texttt{gemma-2-9b-it} \citep{gemma2}, and two accessible via the OpenAI API, \texttt{gpt-3.5-turbo} and \texttt{gpt-4o}. 

\paragraph{Inference Setting} We explore three prompts in this study (summarized in Table~\ref{tab:prompts_main}), and perform greedy decoding. The \zeroshot{} prompt is from prior work~\cite{bigelow2016large}, and we develop two additional prompts. For the \cotfull{} prompt (\cothought{}), we provide two in-context demonstrations of the number and city game with answers that explicitly list a property of the input to arrive at the yes/no decision~\citep{wei2022chain}. This aims to guide models to invoke entity-specific knowledge. Next, we perform more explicit guidance, and prompt it with the attributes of all the examples in the input $\{x_1, \dots, x_n\}$ as well as the query $y_i$ before providing the same task prompt as in \cref{sec:models}. We call this the \knowledge{} prompt. 

We perform minor post-processing of the LM outputs: we lowercase the generated response, and then check for keywords such as ``yes'' and other strings we recorded from a brief skim of the outputs---e.g., ``likely'' or ``i think the program will also produce...''. To make sure that the set of keywords is thorough, we go through all of the responses that did not contain these keywords and check for phrases indicating a "no" or some degree of uncertaintly --- e.g., "impossible to predict". All such responses contain at least one of these phrases. A full list of words and phrases we used is shown in \Cref{sec:keywords}.

\begin{table}
\small
\centering
\begin{tabular}{p{1.8cm}p{11.2cm}}
\toprule
\textbf{Template} & \textbf{Prompt with an example \textcolor{blue}{input} and \textcolor{purple}{query}}\\ \midrule
\zeroshot{} & There is a computer program that produces \{integers (up to 100)/cities\}. Let’s say it produced the following values: \textcolor{blue}{64, 96}. Question: Do you think it will also produce \textcolor{purple}{49}? Answer: \{yes/no\}\\ \midrule 
 \cotfull{}(\cothought{}) & Q: There is a computer program that produces integers between 0 and 100. Let’s say that it produced the following values: 36, 21, 75, 84. Do you think it will also produce 24? A: Yes, because the program seems to be producing integers that are divisible by 3, and 24 is divisible by 3. 

  \colorbox{black!10}{\textit{\{one more similar in-context example\}}}
 
 Q: There is a computer program that produces integers between 0 and 100. Let’s say that it produced the following values: \textcolor{blue}{64, 96}. Do you think it will also produce \textcolor{purple}{49}? Answer: \{yes/no\}\\ \midrule

\knowledge{}   & \textcolor{blue}{64 is even, a perfect square, a perfect cube, ends with 4, is divisible by 4 and 8, and is a power of 2, 4, and 8. 96 is even, ends with 6, and is divisible by 3, 4, 6, 8, and 12.} \textcolor{purple}{49 is odd, a perfect square, ends with 9, is divisible by 7, and is a power of 7.} There is a computer program that produces integers (up to 100). Let’s say it produced the following values: \textcolor{blue}{64, 96}. Do you think it will also produce \textcolor{purple}{49}? Please only provide a "yes" or "no" without outputting anything else.\\ \bottomrule
\end{tabular}
\caption{Example prompt used in our study. The \zeroshot{} prompt is taken from prior work \citep{bigelow2016large}. We propose two new prompts (\cothought{} and \knowledge{}) that encourage LLMs to consider hypothesis spaces. \cothought{} encourages models to make inferences on hypothesis and set membership, while \knowledge{} directly provides membership information of the elements in the input set and the query. }\vspace{-10pt}
\label{tab:prompts_main}
\end{table}

\paragraph{Parametric knowledge of LMs}
\label{sec:knowledge}

To exhibit sensitivity to suspicious coincidences, LMs must be equipped with the knowledge about individual entities in the input sets as well as other members. For example, if a model does not represent San Francisco to be located in Western hemisphere, it will be hard to show systematic sensitivity to the smallest, most specific hypothesis. 

We estimate (1) the models' knowledge on individual entities (e.g.,  numbers' or cities' membership in our target hypotheses), and (2) how well these hypotheses are predicted for the input-examples in our datasets. For the former, we create prompts for each hypothesis---e.g., \texttt{Is 2 an even number?} or \texttt{Is Kinshasa a capital city?} and measure the models' F1 per hypothesis. For the latter, we create prompts for each input set---e.g., \texttt{Are all the numbers in the set [16, 32] divisible by 4?} or \texttt{Are all the cities in the set [Frankfurt, Marseille] in Europe?}, and measure models' F1 per input set. \Cref{tab:knowledge-f1s} shows these results in terms of average F1s, across both domains and types of analyses. We report the individual numbers in Table~\ref{tab:knowledge_numbers} and \ref{tab:knowledge_cities} in the appendix.

\begin{wraptable}{r}{0.6\textwidth}
\centering
\vspace{-1em}
\resizebox{0.6\columnwidth}{!}{
\begin{tabular}{@{}lcccc@{}}
\toprule
\multirow{2}{*}{\textbf{Model}} & \multicolumn{2}{c}{\textbf{Hypothesis}}         & \multicolumn{2}{c}{\textbf{Input}}              \\ \cmidrule(l){2-5} 
           & \textbf{Numbers} & \textbf{Cities} & \textbf{Numbers} & \textbf{Cities} \\ \midrule
Mistral-7B & $0.38_{0.05}$    & $0.81_{0.03}$   & $0.37_{0.02}$    & $0.81_{0.01}$   \\
Llama-3-8B & $0.47_{0.05}$    & $0.81_{0.04}$   & $0.54_{0.02}$    & $0.82_{0.01}$   \\
Gemma-2-9B & $0.68_{0.04}$    & $0.86_{0.02}$   & $0.55_{0.03}$    & $0.82_{0.01}$   \\
GPT-3.5    & $0.71_{0.04}$    & $0.84_{0.03}$   & $0.59_{0.02}$    & $0.83_{0.01}$   \\
GPT-4o                          & $\mathbf{0.98_{0.01}}$ & $\mathbf{0.91_{0.01}}$ & $\mathbf{0.78_{0.02}}$ & $\mathbf{0.87_{0.01}}$ \\ \bottomrule
\end{tabular}
}
\caption{LMs' knowledge results: macro-averaged F1 scores on their knowledge on individual hypotheses (\textbf{Hypothesis}) and on the hypotheses that fit the input sets (\textbf{Input}). Subscripts indicate standard error.}
\label{tab:knowledge-f1s}
\vspace{-2em}
\end{wraptable}

The overall knowledge about Numbers and Cities across LMs is vastly different. While most LMs perform quite well at predicting hypothesis-specific attributes for the cities (all above 81 F1), many models struggle on numbers---e.g., Mistral-7B, which gets an average of 38 F1. This difference carries over to the models' knowledge about the input sets. We will later discuss how this knowledge level impacts LMs' sensitivity to suspicious coincidences.

\section{Results}
\label{sec:exps}
\begin{figure}[t]
    \centering
    \includegraphics[width=0.9\linewidth]{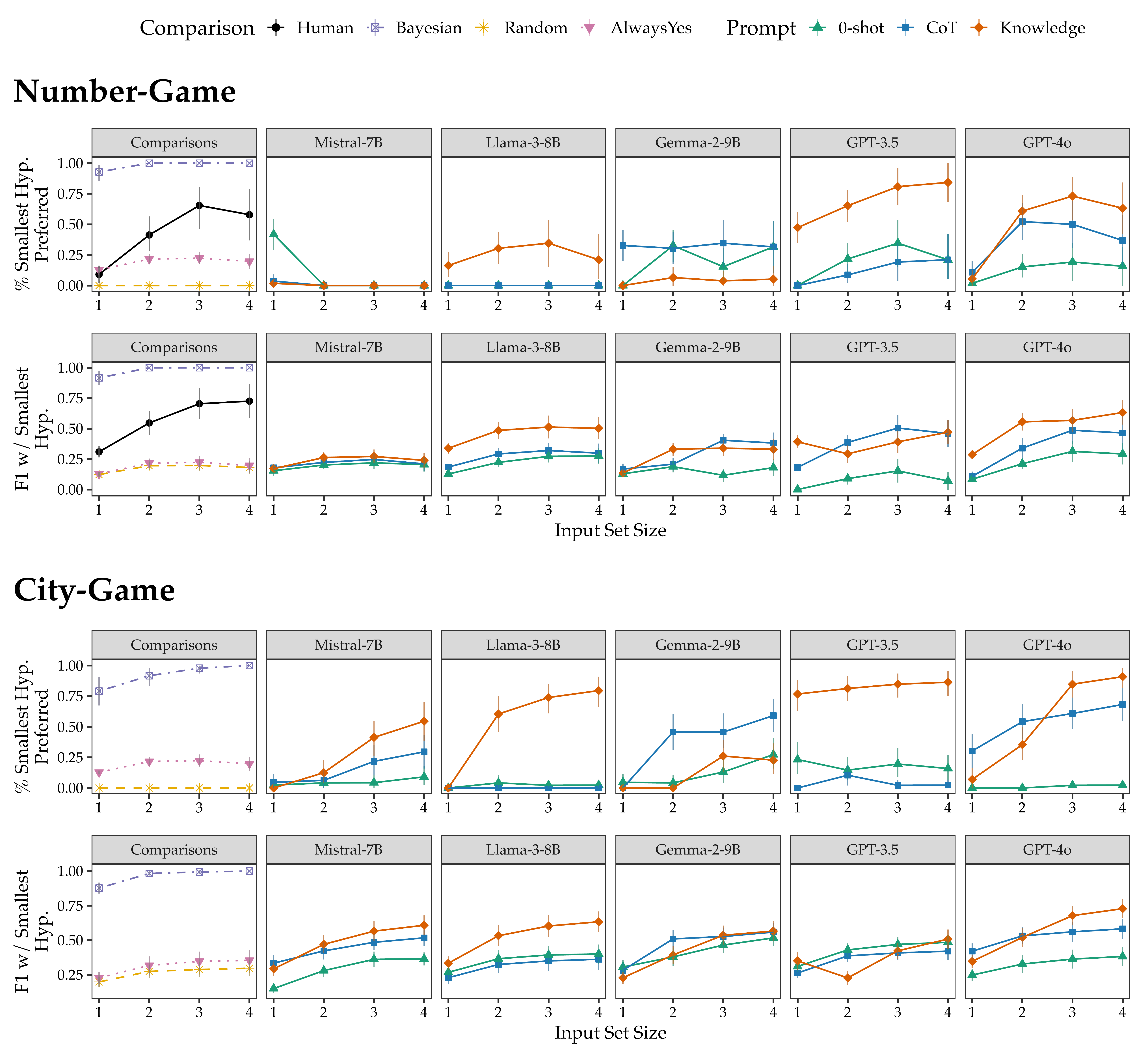}
    \vspace{-1em}
    \caption{Percentage of time models/humans prefer the smallest hypothesis and the Avg. F1 of the smallest hypotheses with the model/human predictions, as a function of input size, for Numbers and Cities. In both cases, an increase in the respective metric as the input size increases is indicative of a suspicious coincidence effect. Baselines and Comparison systems are under the `Comparison' column, and the five LLMs along with the three prompting methods are shown as separate columns. Error bars indicate 95\% confidence intervals.}
    \label{fig:all-choose}
    \vspace{-1em}
\end{figure}

For each system, we report two metrics (the frequency of selecting the smallest hypothesis, and the F1 score between the smallest hypothesis and its predictions). We plot this for two different domains (Number, Cities) separately, and for each plot, we provide results per the size of the input set. We expect both metrics to increase as the input provides more evidence---i.e., the F1 of \hypothesis{pow2} and the predictions for the set $\{16, 32, 2\}$ should be greater than that for the set $\{16\}$. Figure~\ref{fig:all-choose} provides results for all systems over two domains. 

\paragraph{Baseline systems}

In the left-most column, we plot the results from non-LM comparison systems (\textbf{Human, Bayesian model, Random baseline, AlwaysYes baseline}). As expected, the \textbf{Bayesian model} almost always prefers the smallest hypothesis. \textbf{Humans} strongly prefer the smallest hypothesis as the input size increases, though they are not as sensitive as the Bayesian model. Unsurprisingly, the two baselines do not show much of a trend, showing low numbers without any sensitivity to input set size. With these systems, we do not see a meaningful difference between the two metrics (\% smallest hypothesis preferred and the F1 with smallest hypotheses) or two domains. 

\paragraph{Not all LMs are sensitive to suspicious coincidences with the zero-shot prompt.}
Next to the baselines, each plot shows the performance of the same LM with three types of prompts (\zeroshot{}, \cothought{}, \knowledge{}). First, we see that the \zeroshot{} prompt shows low values in both metrics throughout, and mostly does not show an increase in the alignment with the smallest hypothesis as the input set size increases. Overall, LMs with the \zeroshot{} prompt do not score higher than random/AlwaysYes baselines in most cases. Looking a bit deeper into the results, we find that in the \zeroshot{} case, LMs produce `Yes' most of the time---e.g., for the number game, Llama-3-8B produces a `Yes' for 85\% of all queries, while \textbf{Humans} only do this 24\% of the time (see \cref{tab:numbers_yes} and \cref{tab:cities_yes} in the Appendix). A minimal exception to this trend is GPT-4o, which starts off largely similar to AlwaysYes/random (for input sizes 1 and 2) but eventually improves marginally at sizes 3-4. 

\paragraph{Knowledge-rich prompts can enable LMs to show suspicious coincidence effects.} When prompted using the \knowledge{} prompt, most LMs start to show a similar trend as humans: LMs start to strongly prefer the smallest hypothesis as the input set size increases. The best model (GPT-4o) exhibits a behavior consistent with humans on the number game according to both metrics. While GPT-4o already is equipped with the knowledge, as shown in Table~\ref{tab:knowledge-f1s}, such knowledge does not naturally lead the LM to respond to suspicious coincidences in the \zeroshot{} case. The results from other models are less consistent. GPT-3.5, when equipped with the \knowledge{} prompt, favors the smallest hypothesis strongly among all feasible hypotheses, yet without as drastic a gain in its absolute sensitivity (F1 with the smallest hypothesis shows some gains in numbers domain, none in city domain). Llama-3-8B and Mistral-7B, on the other hand, show substantial sensitivity in the city domain but less so in the number domain. With the Gemma-2-9B model, we do not see clear patterns. Overall this result shows that stronger models (i.e., GPT-4o) benefits more from the \knowledge{} prompt, closely matching human behavior.  

\paragraph{CoT prompt enables LMs to favor the smallest hypothesis, when LMs have sufficient knowledge.} Turning to \cothought{}, we observe that on GPT-4o, where parametric knowledge is sufficient, the \cothought{} prompt yields a trend similar to the \knowledge{} prompt. 
It is possible that \cothought{} invokes the knowledge strongly present in GPT-4o, guiding it towards a preference for smaller hypotheses.
In general, \cothought{} seems to occasionally show improvements but is often lower in compatibility with the smallest hypothesis compared to the \knowledge{} setting, sometimes performing no better than \zeroshot{} and random (Llama-3-8B \& Mistral-7B numbers domain). Earlier we observed LMs overall have more knowledge about Cities than Numbers (see Table~\ref{tab:knowledge-f1s}), which might explain the relatively weak performance of \cothought{} on Numbers.

\section{Related Work}
\label{sec:background}

\paragraph{Inductive Reasoning in LMs}

Broader literature investigates inductive reasoning in LMs, specifically over categories \citep{misra2021language, misra2022property, yang2022language, han2024inductive, qiu2024phenomenal, wang2024hypothesis, liu2024incomplete}. While most of these works involve input samples that involve multiple competing hypotheses, many of them deal with entire sets of task examples where the final set fully disambiguates the ``ground-truth'' hypothesis \citep{qiu2024phenomenal, wang2024hypothesis, liu2024incomplete}. As a result, the data used in these works does not enable pragmatically relevant phenomena like suspicious coincidence---which we study using a paradigm that was used in the number game, where subjects are only provided with a handful examples which remain ambiuguous with respect to the hypotheses throughout the experiment.  
Other works, where suspicious coincidence is not necessarily ruled out \citep{misra2022property, han2024inductive}, do not relate LMs' behavior to their knowledge of the hypothesis space (which we look at in sections \ref{sec:knowledge} and \ref{sec:exps}).

\paragraph{Connecting Neural Networks and Bayesian Models} Suspicious coincidences are \textit{naturally} encoded in Bayesian models of cognition, which lie at a different level on Marr's \citeyearpar{marr1982vision} levels of analysis than neural networks (and by extension, LMs). As such, we join recent works drawing connections between the two paradigms \citep[see][for an overview]{griffiths2024bayes}. For instance, \citet{zhu2024eliciting} use the paradigm of ``Gibbs sampling with people'' \citep{harrison2020gibbs} to elicit the priors of LMs. \citet{mccoy2024embers} attempt to use Bayesian reasoning to diagnose cases where LMs tend to fail. \citet{mccoy2023modeling} propose a method to instantiate priors from a Bayesian model as inductive biases for LMs resulting in sample-efficient learning, and \citet{ellis2023human} combines LMs and Bayesian models to propose a hybrid model that correlates strongly with human behavior. While our work does not explicitly combine LMs and Bayesian models, we attempt to guide LMs to operate over hypotheses similar to those utilized by a Bayesian learner of the number (and city) game via prompting. Insofar as this constrains the LMs' computations to be over relevant hypothesis, we further tease apart the access to knowledge vs. the reasoning mechanism between these models.

\section{Discussion and Conclusion}
Humans readily make pragmatically sensitive inductive inferences. This manifests during language comprehension, where listeners make assumptions about the speaker's intended meaning---assuming for instance that speakers choose their words \textit{informatively}. Computationally, informativeness has been translated to the idea that the presented linguistic data has been sampled from the true category/meaning \citep{tenenbaum2001generalization, xu2007sensitivity, frank2014inferring, ransom2016leaping}. This gives rise to the suspicious coincidence effect. In this work, we test the emergence (or lack thereof) of suspicious coincidence effects in LMs, as a means to relate their reasoning behavior to that of humans. This type of nuanced sensitivity in the presence of ambiguity has rarely been studied in the context of LMs' inductive reasoning behavior, with most works focusing on their ability to propose/apply specific rules \citep{ellis2023human, qiu2024phenomenal, wang2024hypothesis, liu2024incomplete}, or make inductive generalizations without necessarily relating them to the relevant hypothesis spaces \citep{misra2022property, han2024inductive}.

Our results suggest that sensitivity to suspicious coincidences can arise in LMs when guided to access the hypothesis landscape. At a high level, this places LMs on level ground with a Bayesian solution---i.e., both models have a more restricted set of candidate hypotheses to operate over (hypothesized in the case of LMs). On the other hand when LMs are only provided with just the task instructions, zero-shot---which is similar to the setup humans are provided with (i.e., no assumptions about what hypotheses they reason over)---they do not demonstrate this sensitivity, instead behaving similar to a random baseline, regardless of how much knowledge about the data are stored in LMs' parameters.

Our findings have several implications: First, LMs' struggling to perform human-like inductive reasoning \citep{qiu2024phenomenal, ellis2023human, han2024inductive} can perhaps be attributed to a lack of explicit access to the relevant hypothesis space. While the inability of LMs to activate relevant hypotheses zero-shot is indeed a limitation, it does not rule out a scenario in which LMs are unable to narrow in on more specific hypotheses when demonstrating inductive reasoning behavior. The emergence of sensitivity to spurious coincidences requires explication of the hypothesis space, similar to how hypotheses are pre-defined in a Bayesian model.
Our findings also motivate methods that can make LMs sensitive to human-like suspicious coincidence effects in a zero-shot manner---e.g., by using distillation to match their output distribution to that of a model that has access to the hypothesis landscape \citep{padmanabhan2023propagating}. Finally, our results provide rich empirical evidence that could be used to further study mechanistic differences between LM states belonging to the same underlying architecture  \citep{geiger2023causal, mueller2024quest, sharkey2025open}, but which differ in their sensitivity to communicative goals.

 \section*{Acknowledgments}
 We thank the members of UT NLP research community for helpful discussion and feedback throughout the project. The project is partially funded by gift from Apple. This work was done in part while the last author was visiting the Simons Institute for the Theory of Computing. 
 
\bibliographystyle{plainnat}
\bibliography{colm2025_conference}

\begin{thebibliography}{37}
\providecommand{\natexlab}[1]{#1}
\providecommand{\url}[1]{\texttt{#1}}
\expandafter\ifx\csname urlstyle\endcsname\relax
  \providecommand{\doi}[1]{doi: #1}\else
  \providecommand{\doi}{doi: \begingroup \urlstyle{rm}\Url}\fi

\bibitem[Bigelow and Piantadosi(2016)]{bigelow2016large}
Eric Bigelow and Steven~T. Piantadosi.
\newblock A large dataset of generalization patterns in the number game.
\newblock \emph{Journal of Open Psychology Data}, 2016.

\bibitem[Ellis(2023)]{ellis2023human}
Kevin Ellis.
\newblock Human-like few-shot learning via bayesian reasoning over natural language.
\newblock \emph{Advances in Neural Information Processing Systems}, 36:\penalty0 13149--13178, 2023.

\bibitem[Frank and Goodman(2014)]{frank2014inferring}
Michael~C Frank and Noah~D Goodman.
\newblock Inferring word meanings by assuming that speakers are informative.
\newblock \emph{Cognitive psychology}, 75:\penalty0 80--96, 2014.

\bibitem[Geiger et~al.(2023)Geiger, Ibeling, Zur, Chaudhary, Chauhan, Huang, Arora, Wu, Goodman, Potts, et~al.]{geiger2023causal}
Atticus Geiger, Duligur Ibeling, Amir Zur, Maheep Chaudhary, Sonakshi Chauhan, Jing Huang, Aryaman Arora, Zhengxuan Wu, Noah Goodman, Christopher Potts, et~al.
\newblock Causal abstraction: A theoretical foundation for mechanistic interpretability.
\newblock \emph{arXiv preprint arXiv:2301.04709}, 2023.

\bibitem[Grattafiori et~al.(2024)Grattafiori, Dubey, Jauhri, Pandey, Kadian, et~al.]{llama3}
Aaron Grattafiori, Abhimanyu Dubey, Abhinav Jauhri, Abhinav Pandey, Abhishek Kadian, et~al.
\newblock The llama 3 herd of models, 2024.
\newblock URL \url{https://arxiv.org/pdf/2407.21783}.

\bibitem[Griffiths et~al.(2024)Griffiths, Zhu, Grant, and Thomas~McCoy]{griffiths2024bayes}
Thomas~L Griffiths, Jian-Qiao Zhu, Erin Grant, and R~Thomas~McCoy.
\newblock Bayes in the age of intelligent machines.
\newblock \emph{Current Directions in Psychological Science}, 33\penalty0 (5):\penalty0 283--291, 2024.

\bibitem[Han et~al.(2024)Han, Ransom, Perfors, and Kemp]{han2024inductive}
Simon~Jerome Han, Keith~J Ransom, Andrew Perfors, and Charles Kemp.
\newblock Inductive reasoning in humans and large language models.
\newblock \emph{Cognitive Systems Research}, 83:\penalty0 101155, 2024.

\bibitem[Harrison et~al.(2020)Harrison, Marjieh, Adolfi, van Rijn, Anglada-Tort, Tchernichovski, Larrouy-Maestri, and Jacoby]{harrison2020gibbs}
Peter Harrison, Raja Marjieh, Federico Adolfi, Pol van Rijn, Manuel Anglada-Tort, Ofer Tchernichovski, Pauline Larrouy-Maestri, and Nori Jacoby.
\newblock Gibbs sampling with people.
\newblock \emph{Advances in neural information processing systems}, 33:\penalty0 10659--10671, 2020.

\bibitem[Jiang et~al.(2023)Jiang, Sablayrolles, Mensch, Bamford, Chaplot, et~al.]{mistral7b}
Albert~Q. Jiang, Alexandre Sablayrolles, Arthur Mensch, Chris Bamford, Devendra Chaplot, et~al.
\newblock Mistral 7b, 2023.
\newblock URL \url{https://arxiv.org/pdf/2310.06825}.

\bibitem[Lake et~al.(2017)Lake, Ullman, Tenenbaum, and Gershman]{lake2017building}
Brenden~M Lake, Tomer~D Ullman, Joshua~B Tenenbaum, and Samuel~J Gershman.
\newblock Building machines that learn and think like people.
\newblock \emph{Behavioral and brain sciences}, 40:\penalty0 e253, 2017.

\bibitem[Leidinger et~al.(2024)Leidinger, Van~Rooij, and Shutova]{leidinger2024llms}
Alina Leidinger, Robert Van~Rooij, and Ekaterina Shutova.
\newblock Are llms classical or nonmonotonic reasoners? lessons from generics.
\newblock In \emph{Proceedings of the 62nd Annual Meeting of the Association for Computational Linguistics (Volume 2: Short Papers)}, pages 558--573, 2024.

\bibitem[Liu et~al.(2024{\natexlab{a}})Liu, Neubig, and Andreas]{liu2024incomplete}
Emmy Liu, Graham Neubig, and Jacob Andreas.
\newblock An incomplete loop: Instruction inference, instruction following, and in-context learning in language models, 2024{\natexlab{a}}.

\bibitem[Liu et~al.(2024{\natexlab{b}})Liu, Min, Zettlemoyer, Choi, and Hajishirzi]{liu2024infinigramscalingunboundedngram}
Jiacheng Liu, Sewon Min, Luke Zettlemoyer, Yejin Choi, and Hannaneh Hajishirzi.
\newblock Infini-gram: Scaling unbounded n-gram language models to a trillion tokens, 2024{\natexlab{b}}.
\newblock URL \url{https://arxiv.org/abs/2401.17377}.

\bibitem[Marr(1982)]{marr1982vision}
David Marr.
\newblock Vision: A computational investigation into the human representation and processing of visual information, 1982.

\bibitem[McCoy and Griffiths(2023)]{mccoy2023modeling}
R~Thomas McCoy and Thomas~L Griffiths.
\newblock Modeling rapid language learning by distilling bayesian priors into artificial neural networks.
\newblock \emph{arXiv preprint arXiv:2305.14701}, 2023.

\bibitem[McCoy et~al.(2024)McCoy, Yao, Friedman, Hardy, and Griffiths]{mccoy2024embers}
R~Thomas McCoy, Shunyu Yao, Dan Friedman, Mathew~D Hardy, and Thomas~L Griffiths.
\newblock Embers of autoregression show how large language models are shaped by the problem they are trained to solve.
\newblock \emph{Proceedings of the National Academy of Sciences}, 121\penalty0 (41):\penalty0 e2322420121, 2024.

\bibitem[Medin et~al.(2003)Medin, Coley, Storms, and Hayes]{medin2003relevance}
Douglas~L Medin, John~D Coley, Gert Storms, and Brett~L Hayes.
\newblock A relevance theory of induction.
\newblock \emph{Psychonomic Bulletin \& Review}, 10\penalty0 (3):\penalty0 517--532, 2003.

\bibitem[Misra et~al.(2021)Misra, Ettinger, and Rayz]{misra2021language}
Kanishka Misra, Allyson Ettinger, and Julia Rayz.
\newblock Do language models learn typicality judgments from text?
\newblock In \emph{Proceedings of the Annual Meeting of the Cognitive Science Society}, volume~43, 2021.

\bibitem[Misra et~al.(2022)Misra, Rayz, and Ettinger]{misra2022property}
Kanishka Misra, Julia Rayz, and Allyson Ettinger.
\newblock A property induction framework for neural language models.
\newblock In \emph{Proceedings of the Annual Meeting of the Cognitive Science Society}, volume~44, 2022.

\bibitem[Mueller et~al.(2024)Mueller, Brinkmann, Li, Marks, Pal, Prakash, Rager, Sankaranarayanan, Sharma, Sun, et~al.]{mueller2024quest}
Aaron Mueller, Jannik Brinkmann, Millicent Li, Samuel Marks, Koyena Pal, Nikhil Prakash, Can Rager, Aruna Sankaranarayanan, Arnab~Sen Sharma, Jiuding Sun, et~al.
\newblock The quest for the right mediator: A history, survey, and theoretical grounding of causal interpretability.
\newblock \emph{arXiv preprint arXiv:2408.01416}, 2024.

\bibitem[Navarro and Perfors(2010)]{navarro2010similarity}
Daniel~J Navarro and Amy~F Perfors.
\newblock Similarity, feature discovery, and the size principle.
\newblock \emph{Acta Psychologica}, 133\penalty0 (3):\penalty0 256--268, 2010.

\bibitem[Osherson et~al.(1990)Osherson, Smith, Wilkie, Lopez, and Shafir]{osherson1990category}
Daniel~N Osherson, Edward~E Smith, Ormond Wilkie, Alejandro Lopez, and Eldar Shafir.
\newblock Category-based induction.
\newblock \emph{Psychological review}, 97\penalty0 (2):\penalty0 185, 1990.

\bibitem[Padmanabhan et~al.(2023)Padmanabhan, Onoe, Zhang, Durrett, and Choi]{padmanabhan2023propagating}
Shankar Padmanabhan, Yasumasa Onoe, Michael Zhang, Greg Durrett, and Eunsol Choi.
\newblock Propagating knowledge updates to lms through distillation.
\newblock \emph{Advances in Neural Information Processing Systems}, 36:\penalty0 47124--47142, 2023.

\bibitem[Qiu et~al.(2024)Qiu, Jiang, Lu, Sclar, Pyatkin, Bhagavatula, Wang, Kim, Choi, Dziri, and Ren]{qiu2024phenomenal}
Linlu Qiu, Liwei Jiang, Ximing Lu, Melanie Sclar, Valentina Pyatkin, Chandra Bhagavatula, Bailin Wang, Yoon Kim, Yejin Choi, Nouha Dziri, and Xiang Ren.
\newblock Phenomenal yet puzzling: Testing inductive reasoning capabilities of language models with hypothesis refinement.
\newblock In \emph{Proceedings of the 12th International Conference on Learning Representations}, 2024.

\bibitem[Ransom et~al.(2016)Ransom, Perfors, and Navarro]{ransom2016leaping}
Keith~J Ransom, Andrew Perfors, and Danielle~J Navarro.
\newblock Leaping to conclusions: Why premise relevance affects argument strength.
\newblock \emph{Cognitive Science}, 40\penalty0 (7):\penalty0 1775--1796, 2016.

\bibitem[Raviv et~al.(2022)Raviv, Lupyan, and Green]{raviv2022variability}
Limor Raviv, Gary Lupyan, and Shawn~C Green.
\newblock How variability shapes learning and generalization.
\newblock \emph{Trends in cognitive sciences}, 26\penalty0 (6):\penalty0 462--483, 2022.

\bibitem[Riviere et~al.(2024)Riviere, Pathak, Sessa, Hardin, Bhupatiraju, et~al.]{gemma2}
Morgane Riviere, Shreya Pathak, Pier Sessa, Cassidy Hardin, Surya Bhupatiraju, et~al.
\newblock Gemma 2: Improving open language models at a practical size, 2024.
\newblock URL \url{http://arxiv.org/pdf/2408.00118}.

\bibitem[Sharkey et~al.(2025)Sharkey, Chughtai, Batson, Lindsey, Wu, Bushnaq, Goldowsky-Dill, Heimersheim, Ortega, Bloom, et~al.]{sharkey2025open}
Lee Sharkey, Bilal Chughtai, Joshua Batson, Jack Lindsey, Jeff Wu, Lucius Bushnaq, Nicholas Goldowsky-Dill, Stefan Heimersheim, Alejandro Ortega, Joseph Bloom, et~al.
\newblock Open problems in mechanistic interpretability.
\newblock \emph{arXiv preprint arXiv:2501.16496}, 2025.

\bibitem[Tenenbaum(1999)]{tenenbaum1999Bayesian}
Joshua Tenenbaum.
\newblock \emph{A Bayesian framework for concept learning}.
\newblock PhD thesis, Massachusetts Institute of Technology, 1999.

\bibitem[Tenenbaum(2000)]{tenenbaum2000rules}
Joshua Tenenbaum.
\newblock Rules and similarity in concept learning.
\newblock \emph{Advances in Neural Information Processing Systems}, 12, 2000.

\bibitem[Tenenbaum and Griffiths(2001)]{tenenbaum2001generalization}
Joshua~B Tenenbaum and Thomas~L Griffiths.
\newblock Generalization, similarity, and bayesian inference.
\newblock \emph{Behavioral and brain sciences}, 24\penalty0 (4):\penalty0 629--640, 2001.

\bibitem[Wang et~al.(2024)Wang, Zelikman, Poesia, Pu, Haber, and Goodman]{wang2024hypothesis}
Roucheng Wang, Eric Zelikman, Gabriel Poesia, Yewen Pu, Nick Haber, and Noah~D. Goodman.
\newblock Hypothesis search: Inductive reasoning with language models.
\newblock In \emph{Proceedings of the 12th International Conference on Learning Representations}, 2024.

\bibitem[Wei et~al.(2022)Wei, Wang, Schuurmans, Bosma, Xia, Chi, Le, Zhou, et~al.]{wei2022chain}
Jason Wei, Xuezhi Wang, Dale Schuurmans, Maarten Bosma, Fei Xia, Ed~Chi, Quoc~V Le, Denny Zhou, et~al.
\newblock Chain-of-thought prompting elicits reasoning in large language models.
\newblock \emph{Advances in neural information processing systems}, 35:\penalty0 24824--24837, 2022.

\bibitem[Xu and Tenenbaum(2007{\natexlab{a}})]{xu2007sensitivity}
Fei Xu and Joshua~B Tenenbaum.
\newblock Sensitivity to sampling in bayesian word learning.
\newblock \emph{Developmental science}, 10\penalty0 (3):\penalty0 288--297, 2007{\natexlab{a}}.

\bibitem[Xu and Tenenbaum(2007{\natexlab{b}})]{xu2007word}
Fei Xu and Joshua~B Tenenbaum.
\newblock Word learning as bayesian inference.
\newblock \emph{Psychological review}, 114\penalty0 (2):\penalty0 245, 2007{\natexlab{b}}.

\bibitem[Yang et~al.(2022)Yang, Dong, Du, Cheng, Cambria, Liu, Gao, and Wei]{yang2022language}
Zonglin Yang, Li~Dong, Xinya Du, Hao Cheng, Erik Cambria, Xiaodong Liu, Jianfeng Gao, and Furu Wei.
\newblock Language models as inductive reasoners.
\newblock \emph{arXiv preprint arXiv:2212.10923}, 2022.

\bibitem[Zhu and Griffiths(2024)]{zhu2024eliciting}
Jian-Qiao Zhu and Thomas~L Griffiths.
\newblock Eliciting the priors of large language models using iterated in-context learning.
\newblock \emph{arXiv preprint arXiv:2406.01860}, 2024.

\end{thebibliography}

\appendix
\section{Number and City game hypotheses}
\label{appendix:hypotheses}

\paragraph{Number Game Hypotheses} \Cref{tab:number-hyps} describes the set of all hypotheses in our hypothesis space, the number of elements (and some example elements) in the range $1-100$ that fits each hypothesis, and the number of input sets in our set of inputs that fit each hypothesis.

\paragraph{City Game Hypotheses} \Cref{tab:city-hyps} describes the set of all hypotheses in our hypothesis space, the number of elements (and some example elements) in our set of cities that fits each hypothesis, and the number of input sets in our set of inputs that fit each hypothesis.

\begin{table}[h]
\centering
\begin{tabular}{@{}llrr@{}}
\toprule
\textbf{Hypothesis}              & \textbf{Extension}      & \textbf{Size} & \textbf{Freq. in Data} \\ \midrule
\texttt{odd}    & \{1, 3, 5, ..., 99\}    & 50            & 75                     \\
\texttt{even}   & \{2, 4, 6, ..., 100\}   & 50            & 68                     \\
\texttt{square} & \{1, 4, 9, ..., 100\}   & 10            & 13                     \\
\texttt{cube}   & \{1, 8, 27, 64\}        & 4             & 4                      \\
\texttt{prime}  & \{2, 3, 5, ..., 97\}    & 25            & 28                     \\
\texttt{end0}   & \{10, 20, ..., 100\}    & 10            & 7                      \\
\texttt{end1}   & \{11, 21, ..., 91\}     & 10            & 11                     \\
\texttt{end2}   & \{12, 22, ..., 92\}     & 10            & 8                      \\
\texttt{end3}   & \{13, 23, ..., 93\}     & 10            & 15                     \\
\texttt{end4}   & \{14, 24, ..., 94\}     & 10            & 10                     \\
\texttt{end5}   & \{15, 25, ..., 95\}     & 10            & 11                     \\
\texttt{end6}   & \{16, 26, ..., 96\}     & 10            & 11                     \\
\texttt{end7}   & \{17, 27, ..., 97\}     & 10            & 8                      \\
\texttt{end8}   & \{18, 28, ..., 98\}     & 10            & 8                      \\
\texttt{end9}   & \{19, 29, ..., 99\}     & 10            & 8                      \\
\texttt{div3}   & \{3, 6, 9, ..., 99\}    & 33            & 42                     \\
\texttt{div4}   & \{4, 8, 12, ..., 100\}  & 25            & 31                     \\
\texttt{div5}   & \{5, 10, 15, ..., 100\} & 20            & 19                     \\
\texttt{div6}   & \{6, 12, 18, ..., 96\}  & 16            & 16                     \\
\texttt{div7}   & \{7, 14, 21, ..., 98\}  & 14            & 9                      \\
\texttt{div8}   & \{8, 16, 24, ..., 96\}  & 12            & 12                     \\
\texttt{div9}   & \{9, 18, 27, ..., 99\}  & 11            & 10                     \\
\texttt{div11}  & \{11, 22, ..., 99\}     & 9             & 6                      \\
\texttt{div12}  & \{12, 24, ..., 96\}     & 8             & 6                      \\
\texttt{pow2}   & \{1, 2, 4, ..., 64\}    & 7             & 14                     \\
\texttt{pow3}   & \{1, 3, 9, 27, 81\}     & 5             & 6                      \\
\texttt{pow4}   & \{1, 4, 16, 64\}        & 4             & 5                      \\
\texttt{pow5}   & \{1, 5, 25\}            & 3             & 3                      \\
\texttt{pow6}   & \{1, 6, 36\}            & 3             & 3                      \\
\texttt{pow7}   & \{1, 7, 49\}            & 3             & 3                      \\
\texttt{pow8}   & \{1, 8, 64\}            & 3             & 4                      \\
\texttt{pow9}   & \{1, 9, 81\}            & 3             & 4                      \\
\texttt{pow10}  & \{1, 10, 100\}          & 3             & 3                      \\ \bottomrule
\end{tabular}
\caption{The full hypothesis space for the number game experiments, including the number of elements from $1-100$ in its extension and the number of sets in our dataset that fit the hypothesis}
\label{tab:number-hyps}
\end{table}

\begin{table}[h]
\centering
\resizebox{\textwidth}{!}{
\begin{tabular}{@{}llrr@{}}
\toprule
\textbf{Hypothesis} & \textbf{Extension} & \textbf{Size} & \textbf{Freq. in Data} \\ \midrule
Eastern Hemisphere - \texttt{east} & \{Tokyo, Jakarta, Cairo, ...\} & 255 & 85\\
Western Hemisphere - \texttt{west} & \{Sao Paulo, Mexico City, Montreal, ...\} & 245 & 49\\
Northern Hemisphere - \texttt{north} & \{Tokyo, Delhi, New York, ...\} & 449 & 125\\
Southern Hemisphere - \texttt{south} & \{Jakarta, Sao Paulo, Sydney, ...\} & 51 & 36\\
Tropical Latitudes - \texttt{tropical} & \{Jakarta, Manila, Bangkok, ...\} & 108 & 30\\
Temperate Latitudes - \texttt{temperate} & \{Tokyo, Seoul, Buenos Aires, ...\} & 392 & 121\\
Africa - \texttt{africa} & \{Cairo, Lagos, Kinshasa, ...\} & 59 & 19\\
Asia - \texttt{asia} & \{Tokyo, Jakarta, Delhi, ...\} & 111 & 25\\
Europe - \texttt{europe} & \{Moscow, London, Paris, ...\} & 116 & 30\\
North America - \texttt{north america} & \{Mexico City, New York, Los Angeles, ...\} & 178 & 32\\
Oceania - \texttt{oceania} & \{Melbourne, Auckland, Port Moresby, ...\} & 12 & 9\\
South America - \texttt{south america} & \{Sao Paulo, Buenos Aires,  Rio de Janeiro, ...\} & 24 & 10\\
Cities with 10M+ people - \texttt{10M+} & \{Tokyo, Sao Paulo, New York, ...\} & 41 & 31\\
Cities in Developed Countries - \texttt{developed} & \{Tokyo, New York, London, ...\} & 284 & 79\\
Cities in Developing Countries - \texttt{developing} & \{Jakarta, Delhi, Guadalajara, ...\} & 180 & 53\\
Cities in Least-Developed Countries - \texttt{least developed} & \{Dhaka, Kinshasa, Khartoum, ...\} & 36 & 9\\
Capital Cities - \texttt{capital} & \{Tokyo, Seoul, Cairo, ...\} & 144 & 34\\
NFL Host Cities - \texttt{NFL} & \{Los Angeles, Philadelphia, Kansas City, ...\} & 27 & 15\\ \bottomrule
\end{tabular}
}
\caption{The full hypothesis space for the number game experiments, including the number of elements in our set of cities in the extension of each hypothesis and the number of sets in our dataset that fit the hypothesis}
\label{tab:city-hyps}
\end{table}

\section{Fine-grained results on models' knowledge of hypotheses}

\Cref{tab:knowledge_numbers} describes the amount of knowledge the models have of each hypothesis in the numbers domain.
\Cref{tab:knowledge_cities} describes the amount of knowledge the models have of each hypothesis in the cities domain.

\begin{table}[h]
    \centering
    \begin{tabular}{@{}llllll@{}}
    \toprule
          Hypothesis & \texttt{mistral-7b} & \texttt{llama-8b} & \texttt{gemma-9b} & \texttt{gpt-3.5} & \texttt{gpt-4o}\\ \midrule
         odd & $0.67$ & $0.87$ & $0.92$ & $0.72$ & $1.00$\\
         even & $0.30$ & $0.78$ & $0.93$ & $0.81$ & $1.00$\\
         square & $0.50$ & $0.00$ & $0.87$ & $0.51$ & $1.00$\\
         cube & $0.00$ & $0.00$ & $0.57$ & $0.00$ & $1.00$\\
         prime & $0.39$ & $0.51$ & $0.93$ & $0.46$ & $1.00$\\
         ends with $0$ & $0.18$ & $0.76$ & $0.82$ & $1.00$ & $1.00$\\
         ends with $1$ & $0.43$ & $0.71$ & $0.95$ & $0.83$ & $1.00$\\
         ends with $2$ & $0.43$ & $0.50$ & $0.77$ & $0.71$ & $1.00$\\
         ends with $3$ & $0.80$ & $0.61$ & $0.83$ & $0.77$ & $1.00$\\
         ends with $4$ & $0.70$ & $0.52$ & $0.67$ & $0.95$ & $1.00$\\
         ends with $5$ & $0.74$ & $0.44$ & $0.91$ & $0.91$ & $1.00$\\
         
         ends with $6$ & $0.58$ & $0.46$ & $0.71$ & $0.77$ & $1.00$\\
         
         ends with $7$ & $0.59$ & $0.67$ & $0.69$ & $0.83$ & $1.00$\\
         
         ends with $8$ & $0.74$ & $0.45$ & $0.65$ & $0.80$ & $1.00$\\
         
         ends with $9$ & $0.59$ & $0.52$ & $0.71$ & $0.83$ & $1.00$\\
         
         divisible by $3$ & $0.54$ & $0.66$ & $0.96$ & $0.78$ & $0.99$\\
         
         divisible by $4$ & $0.42$ & $0.83$ & $0.81$ & $0.75$ & $0.96$\\
         
         divisible by $5$ & $0.40$ & $0.67$ & $0.97$ & $0.95$ & $1.00$\\
         
         divisible by $6$ & $0.67$ & $0.55$ & $0.68$ & $0.84$ & $0.97$\\
         
         divisible by $7$ & $0.13$ & $0.17$ & $0.78$ & $0.75$ & $1.00$\\
         
         divisible by $8$ & $0.44$ & $0.85$ & $0.65$ & $0.80$ & $0.92$\\
         
         divisible by $9$ & $0.43$ & $0.43$ & $0.49$ & $0.71$ & $1.00$\\
         
         divisible by $11$ & $0.00$ & $0.13$ & $0.55$ & $0.72$ & $0.95$\\
         
         divisible by $12$ & $0.56$ & $0.67$ & $0.62$ & $0.94$ & $1.00$\\
         
         power of $2$ & $0.44$ & $0.93$ & $0.78$ & $0.60$ & $1.00$\\
         
         power of $3$ & $0.00$ & $0.33$ & $0.44$ & $0.50$ & $1.00$\\
         
         power of $4$ & $0.00$ & $0.29$ & $0.32$ & $0.40$ & $1.00$\\
         
         power of $5$ & $0.00$ & $0.00$ & $0.57$ & $0.50$ & $1.00$\\
         
         power of $6$ & $0.00$ & $0.00$ & $0.31$ & $0.80$ & $0.80$\\
         
         power of $7$ & $0.00$ & $0.00$ & $0.25$ & $0.50$ & $1.00$\\
         
         power of $8$ & $0.00$ & $0.00$ & $0.27$ & $0.80$ & $0.80$\\
         
         power of $9$ & $0.00$ & $0.50$ & $0.27$ & $0.57$ & $1.00$\\
         
         power of $10$ & $1.00$ & $0.80$ & $0.80$ & $0.50$ & $0.80$\\ \bottomrule

    \end{tabular}
    \caption{Knowledge F1 Scores for each model for each hypothesis in number domain.}
    \label{tab:knowledge_numbers}
\end{table}

\begin{table}[h]
\centering
\begin{tabular}{@{}llllll@{}}
\toprule
 Hypothesis & \texttt{llama-8b} & \texttt{mistral-7b} & \texttt{gemma-9b} & \texttt{gpt-3.5} & \texttt{gpt-4o} \\ \midrule
East            & 0.854    & 0.860      & 0.803    & 0.885   & 0.892  \\
West            & 0.869    & 0.831      & 0.886    & 0.897   & 0.875  \\
North           & 0.956    & 0.957      & 0.972    & 0.961   & 0.985  \\
South           & 0.817    & 0.623      & 0.816    & 0.777   & 0.933  \\
Tropical        & 0.706    & 0.787      & 0.750    & 0.785   & 0.874  \\
Temperate       & 0.819    & 0.882      & 0.889    & 0.931   & 0.880  \\
Africa          & 0.959    & 0.950      & 0.975    & 0.951   & 0.983  \\
Asia            & 0.897    & 0.915      & 0.945    & 0.945   & 0.950  \\
Europe          & 0.941    & 0.910      & 0.920    & 0.925   & 0.964  \\
North America   & 0.952    & 0.941      & 0.932    & 0.948   & 0.954  \\
Oceania         & 0.917    & 0.917      & 0.880    & 0.880   & 0.880  \\
South America   & 0.844    & 0.844      & 0.851    & 0.880   & 0.920  \\
10M+            & 0.795    & 0.698      & 0.757    & 0.816   & 0.865  \\
Developed       & 0.779    & 0.875      & 0.849    & 0.829   & 0.899  \\
Developing      & 0.430    & 0.622      & 0.802    & 0.610   & 0.846  \\
Least Developed & 0.440    & 0.667      & 0.642    & 0.719   & 0.854  \\
Capital         & 0.880    & 0.603      & 0.929    & 0.891   & 0.951  \\
NFL             & 0.761    & 0.745      & 0.812    & 0.475   & 0.892 \\ \bottomrule
\end{tabular}
    \caption{Knowledge F1 Scores for each model for each hypothesis in city domain.}
    \label{tab:knowledge_cities}
\end{table}

\section{Keywords used to get LM responses}
\label{sec:keywords}
We use the following keywords/strategy to measure and track LM responses, all after the outputs are lowercased:
\begin{quote}
\texttt{``yes'', ``likely'' (made sure ``unlikely'' wasn't in output), ``i think it will also produce'', ``i think the program will also produce'', ``i predict that it will also produce''}
\end{quote}

\section{Prompts}

\Cref{tab:number_prompts} gives example Chain-of-Thought and Knowledge-Rich prompts provided to the models for the numbers domain.

\Cref{tab:city_prompts} gives example Chain-of-Thought and Knowledge-Rich prompts provided to the models for the cities domain.

\begin{table}[]
\centering
\begin{tabular}{@{}ll@{}}
\toprule
\textbf{Prompt Type} & \textbf{Example Prompt}\\ \midrule
Chain-of-Thought & \begin{tabular}[c]{@{}l@{}}\texttt{Q: There is a computer program that} \\ \texttt{produces integers between} \\ \texttt{0 and 100. Let’s say that it produced} \\ \texttt{the following values:} \\ \texttt{36, 21, 75, 84. Do you think it will} \\ \texttt{also produce 24?} \\ \\ \texttt{A: Yes, because the program seems to} \\ \texttt{be producing integers that} \\ \texttt{are divisible by 3, and 24 is divisible} \\ \texttt{by 3.} \\ \\ \texttt{Q: There is a computer program that} \\ \texttt{produces integers between} \\ \texttt{0 and 100. Let’s say it produced the} \\ \texttt{following values:} \\ \texttt{3, 17, 25, 39. Do you think it will also} \\ \texttt{produce 24?} \\ \\ \texttt{A: No, because the program seems to} \\ \texttt{be producing odd integers, and 24 is} \\ \texttt{not odd.} \\ \\ \texttt{Q: There is a computer program that} \\ \texttt{produces integers between 0 and 100.} \\ \texttt{Let’s say it produced the following} \\ \texttt{values: 64, 96. Do you think it will} \\ \texttt{also produce 49?} \\ \\ \texttt{A: \{yes/no\}}\end{tabular} \\
\hline
Knowledge-Rich   & \begin{tabular}[c]{@{}l@{}}
\\ \texttt{64 is even, a perfect square,} \\ \texttt{a perfect cube, ends with 4,} \\ \texttt{is divisible by 4 and 8, and is} \\ \texttt{a power of 2, 4, and 8. 96 is} \\ \texttt{even, ends with 6, and is divisible} \\ \texttt{by 3, 4, 6, 8, and 12. 49 is odd, a} \\ \texttt{perfect square, ends with 9, is} \\ \texttt{divisible by 7, and is a power of 7.} \\ \texttt{There is a computer program that} \\ \texttt{produces integers (up to 100).} \\ \texttt{Let's say it produced the following} \\ \texttt{values: 64, 96. Do you think it will} \\ \texttt{also produce 49? Please only} \\ \texttt{provide a "yes" or "no" without} \\ \texttt{outputting anything else.}\\
\\\end{tabular}\\ \bottomrule
\end{tabular}
\caption{Example Chain-of-Thought and Knowledge-Rich prompts for the numbers domain}
\label{tab:number_prompts}
\end{table}

\begin{table}[htbp]
\centering
\resizebox{!}{0.45\textheight}{
\begin{tabular}{@{}ll@{}}
\toprule
\textbf{Prompt Type} & \textbf{Example Prompt}\\ \midrule
Chain-of-Thought & \begin{tabular}[c]{@{}l@{}}\texttt{Q: There is a computer program that} \\ \texttt{produces cities. Let's say it produced} \\ \texttt{the following cities: Paris, Nairobi,} \\ \texttt{Lima, Canberra. Do you think it will} \\ \texttt{also produce Ottawa?}\\ \\ \texttt{A: Yes, because the computer program} \\ \texttt{appears to be generating national} \\ \texttt{capitals, and Ottawa is a national} \\ \texttt{capital.}\\ \\ \texttt{Q: There is a computer program that} \\ \texttt{produces cities. Let's say it produced} \\ \texttt{the following cities: Beijing, Riyadh,} \\ \texttt{Karachi, Phnom Penh. Do you think} \\ \texttt{it will also produce Madrid?}\\ \\ \texttt{A: No, because the computer program} \\ \texttt{appears to be generating cities in Asia,} \\ \texttt{and Madrid is in Europe.}\\ \\ \texttt{Q: There is a computer program that} \\ \texttt{produces cities. Let's say it produced} \\ \texttt{the following cities: Houston,} \\ \texttt{Adelaide, Paris, Busan. Do you think} \\ \texttt{it will also produce Dhaka?}\\ \\ \texttt{A: No, because the computer program} \\ \texttt{appears to be generating cities in} \\ \texttt{developed countries, and Dhaka} \\ \texttt{is in Bangladesh, which is not a} \\ \texttt{developed country.}\\ \\ \texttt{Q: There is a computer program that} \\ \texttt{produces cities. Let's say it produced} \\ \texttt{the following cities: Dubai, Karachi.} \\ \texttt{Do you think it will also produce} \\ \texttt{Tokyo?}\\ \\ \texttt{A: \{yes/no\}}\end{tabular} \\ \\
\hline
Knowledge-Rich & \begin{tabular}[c]{@{}l@{}}\\ \texttt{Dubai is in the Eastern and Northern} \\ \texttt{Hemispheres, is} \\ \texttt{located within temperate latitudes, is} \\ \texttt{in Asia, and is in a developing country.} \\ \texttt{Guangzhou is in the Eastern and} \\ \texttt{Northern Hemispheres, is located}\\ \texttt{within tropical latitudes, is in Asia, has} \\ \texttt{at least ten million people, and is in a} \\ \texttt{developing country. There is a} \\ \texttt{computer program that produces cities.} \\ \texttt{Let’s say it produced the following cities:} \\ \texttt{Dubai. Do you think it will also produce} \\ \texttt{Guangzhou? Please only provide a} \\ \texttt{"yes" or "no" without outputting} \\ \texttt{anything else.}\end{tabular}\\ \\ \bottomrule
\end{tabular}
}
\caption{Example Chain-of-Thought and Knowledge-Rich prompts for the cities domain}
\label{tab:city_prompts}
\end{table}

\section{How often do Models respond with "yes" to a set-target pair?}

\Cref{tab:numbers_yes} and \Cref{tab:cities_yes} show how often systems/humans answer questions with `Yes' for the number and city domains, respectively.

\begin{table}[h]
\centering
\begin{tabular}{lll}
\toprule
\textbf{System}             & \textbf{Prompt} & \textbf{\% Yes} \\ \midrule
Random & - & 57\%\\
Bayesian & - & 12\%\\
Humans & - & 24\%\\ \midrule
\multirow{3}{*}{Mistral-7B} & 0-shot & 52\%\\
                            & CoT & 78\%\\
                            & Knowledge-Rich & 79\%\\ \midrule
\multirow{3}{*}{Llama-8B}   & 0-shot & 85\%\\
                            & CoT & 59\%\\
                            & Knowledge-Rich & 23\%\\ \midrule
\multirow{3}{*}{Gemma-9B}   & 0-shot & 42\%\\
                            & CoT & 12\%\\
                            & Knowledge-Rich & 68\%\\ \midrule
\multirow{3}{*}{GPT-3.5}    & 0-shot & 4\%\\
                            & CoT & 38\%\\
                            & Knowledge-Rich & 7\%\\ \midrule
\multirow{3}{*}{GPT-4o}     & 0-shot & 34\%\\
                            & CoT & 11\%\\
                            & Knowledge-Rich & 25\%\\ \bottomrule          
\end{tabular}
    \caption{The percentage of set-target pairs each model said "yes" to in the numbers domain}
    \label{tab:numbers_yes}
\end{table}

\begin{table}[h]
\centering
\begin{tabular}{lll}
\toprule
\textbf{System}             & \textbf{Prompt} & \textbf{\% Yes} \\ \midrule
Random & - & 57\%\\
Bayesian & - & 23\%\\ \midrule
\multirow{3}{*}{Mistral-7B} & 0-shot & 40\%\\
                            & CoT & 46\%\\
                            & Knowledge-Rich & 48\%\\ \midrule
\multirow{3}{*}{Llama-8B}   & 0-shot & 76\%\\
                            & CoT & 98\%\\
                            & Knowledge-Rich & 30\%\\ \midrule
\multirow{3}{*}{Gemma-9B}   & 0-shot & 56\%\\
                            & CoT & 44\%\\
                            & Knowledge-Rich & 67\%\\ \midrule
\multirow{3}{*}{GPT-3.5}    & 0-shot & 42\%\\
                            & CoT & 51\%\\
                            & Knowledge-Rich & 6\%\\ \midrule
\multirow{3}{*}{GPT-4o}     & 0-shot & 85\%\\
                            & CoT & 21\%\\
                            & Knowledge-Rich & 30\%\\ \bottomrule          
\end{tabular}
    \caption{The percentage of set-target pairs each model said "yes" to in the cities domain}
    \label{tab:cities_yes}
\end{table}

\end{document}